\documentclass{article}

\usepackage{arxiv}

\usepackage[utf8]{inputenc} 
\usepackage[T1]{fontenc}    
\usepackage{hyperref}       
\usepackage{url}            
\usepackage{booktabs}       
\usepackage{amsfonts}       
\usepackage{nicefrac}       
\usepackage{microtype}      
\usepackage{graphicx}
\usepackage{natbib}
\usepackage{doi}
\usepackage{amsmath}

\title{GeoFormer: A Vision and Sequence Transformer-based Approach for Greenhouse Gas Monitoring}

\author{Madhav Khirwar\\
Bangalore, India \\
\texttt{madhavkhirwar49@gmail.com} \\
\And
Ankur Narang \\
New Delhi, India \\
\texttt{ankur.narang@fermionai.com} \\
}

\date{}

\renewcommand{\headeright}{}
\renewcommand{\shorttitle}{}

\hypersetup{
pdftitle={A template for the arxiv style},
pdfsubject={q-bio.NC, q-bio.QM},
pdfauthor={David S.~Hippocampus, Elias D.~Striatum},
pdfkeywords={First keyword, Second keyword, More},
}

\begin{document}

\maketitle

\begin{abstract}

Air pollution represents a pivotal environmental challenge globally, playing a major role in climate change via greenhouse gas emissions and negatively affecting the health of billions. However predicting the spatial and temporal patterns of pollutants remains challenging. The scarcity of ground-based monitoring facilities and the dependency of air pollution modeling on comprehensive datasets, often inaccessible for numerous areas, complicate this issue. In this work, we introduce GeoFormer, a compact model that combines a vision transformer module with a highly efficient time-series transformer module to predict surface-level nitrogen dioxide (NO\textsubscript{2}) concentrations from Sentinel-5P satellite imagery. We train the proposed model to predict surface-level NO\textsubscript{2} measurements using a dataset we constructed with Sentinel-5P images of ground-level monitoring stations, and their corresponding NO\textsubscript{2} concentration readings. The proposed model attains high accuracy (MAE 5.65), demonstrating the efficacy of combining  vision and time-series transformer architectures to harness satellite-derived data for enhanced GHG emission insights, proving instrumental in advancing climate change monitoring and emission regulation efforts globally.
\end{abstract}

\section{Introduction}

The emission of greenhouse gases (GHGs), primarily from industrial and transportation activities, is a major contributor to the increasingly urgent climate change crisis. This article introduces innovative methodologies for forecasting the levels of nitrogen dioxide (NO\textsubscript{2}), a prevalent byproduct of fossil fuel combustion that poses significant risks to both human health and the environment. Notably, NO\textsubscript{2} is closely associated with other air contaminants, such as fine particulate matter (PM2.5), and is often released alongside CO\textsubscript{2}, a leading greenhouse gas, rendering it an effective indicator for gauging CO\textsubscript{2} emissions. The adverse effects of NO\textsubscript{2} on human health, particularly on the cardiovascular and respiratory systems, underscore the necessity of managing NO\textsubscript{2} levels. The ability to accurately identify instances where safe exposure thresholds are exceeded, as well as to assess individual exposure levels, demands detailed insight into the spatial and temporal distribution of NO\textsubscript{2}. This requirement serves as a key motivation for our research.

The advent of high-resolution satellite imagery, such as that provided by the Sentinel-5P satellite, offers unprecedented opportunities for the monitoring of atmospheric pollutants. The TROPOMI device on the Sentinel-5P satellite enables detailed observation of NO\textsubscript{2} emissions on a global scale \cite{bodah2022sentinel}. However, the challenge lies in effectively analyzing this vast amount of data to produce accurate and timely predictions of surface-level GHG concentrations. Although deep learning models have demonstrated potential for estimating GHG emissions with the use of satellite imagery, there is an increasing need for models that are both accurate and computationally efficient. In response to these challenges, this paper presents an innovative architecture based on both a vision transformer (ViT) and a time-series transformer, aimed at monitoring greenhouse gas (GHG) emissions using Sentinel-5P imagery. Being a fraction of the size of models proposed for similar tasks, this represents an advancement in the application of deep learning to environmental monitoring, establishing a new standard for tracking GHG emissions in real-time and with high efficiency and low compute. In addition, it lays the groundwork for future efforts aimed at mitigating climate change.

The following are the main contributions of this work: 
\begin{enumerate}
    \item \textbf{A Dataset of Paired Sentinel-5P and NO\textsubscript{2} Data:} We introduce a comprehensive dataset that pairs high-resolution Sentinel-5P satellite imagery with corresponding surface-level NO\textsubscript{2} concentration measurements. This dataset is characterized by its daily granularity over the course of 15 months.
    
    \item \textbf{A Compact and Efficient Spatio-temporal Transformer Model:} We propose a novel transformer-based model that leverages the spatial and temporal dynamics of NO\textsubscript{2} emissions; the result is a model that outperforms existing methods in terms of accuracy while being significantly more efficient in terms of computational resources required.
\end{enumerate}
\section{Related Work}

\subsection{Transformer Models}

The transformer architecture, introduced by \citet{vaswani2017attention}, revolutionized sequence modeling. The core mechanism here is self-attention, which allows the model to weigh the importance of different parts of the input data relative to each other. Mathematically, the self-attention mechanism can be described as:

\begin{equation}
\text{Attention}(Q, K, V) = \text{softmax}\left(\frac{QK^T}{\sqrt{d_k}}\right)V
\label{equation:attention}
\end{equation}

where $mQ$, $mK$, and $mV$ represent the query, key, and value matrices, respectively, derived from the input, and $d_k$ is the dimensionality of the keys and queries, which serves as a scaling factor. This ability to handle sequences in their entirety parallelizes computation and has led to state-of-the-art results in various sequence modelling tasks \citep{devlin2019bert}. However, as is noted by \citet{zhou2021informer}, the vanilla transformer architecture has $O(N^2)$ time complexity, where $N$ is the number of input tokens.

\subsection{Deep Learning for Greenhouse Gas Emissions}
Historically, the estimation of the spatial distribution of atmospheric pollutants, including greenhouse gases, has predominantly been based on discrete point measurements from specific locales. These measurements are then extrapolated over larger areas through geostatistical techniques such as kriging \cite{kriging} or Land-Use Regression (LUR) models \cite{hoek2008}. Although these traditional methods have proven useful, they are hampered by their need for detailed variable selection and lack the capability for efficient large-scale application.

In contrast, recent progress in the field of deep learning has shown promise for improving the accuracy of greenhouse gas quantification. For instance, \citet{deepNO2, multimodalCO2} leverage Sentinel-2 and Sentinel-5P data to estimate CO\textsubscript{2}and NO\textsubscript{2} surface-level emissions using convolutional neural network (CNN) backbones. \citet{khirwar2023geovit} propose a vision transformer-based approach, achieving better results with a more compact model on the dataset proposed by \citet{deepNO2}. However, this dataset has at most a monthly granularity. Although this may have applications for identifying longer-term trends, such a dataset does not offer the opportunity to detect anomalous spikes in greenhouse gases/pollutants quickly.

\begin{figure}[h]
\begin{center}
\includegraphics[width=\columnwidth]{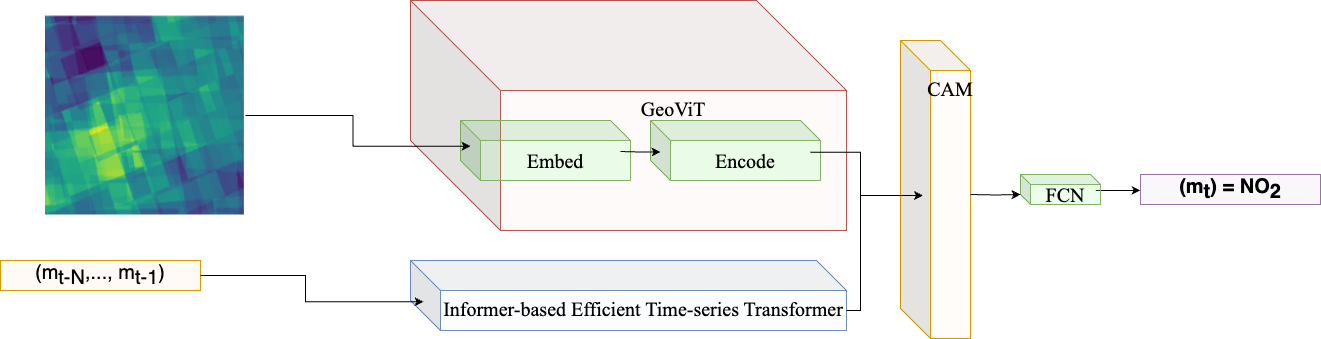}
\end{center}
\caption{GeoFormer model architecture. Here, $m_t$ represents an NO\textsubscript{2} prediction at timestamp $t$, and CAM represents the cross-attention module.}
\label{figure:diagram}
\end{figure}

\section{Methodology}

\subsection{Vision Transformer Module}
Transformers have been adapted to image data by treating images as sequences of patches \cite{dosovitskiy2020}. The fundamental principle here is to decompose an image into a sequence of smaller, fixed-size patches, apply linear transformations to project these patches into an embedding space, and then process the embeddings using the Transformer's self-attention mechanism. This allows each element to attend over all positions, as opposed to convolutional neural networks (CNNs) that are limited by their kernel sizes \cite{wang2021pyramid}. Mathematically,

\begin{equation}
\text{MHSA}(E) = \text{softmax}\left(\frac{E Q E^T K}{\sqrt{d_k}}\right)E V
\end{equation}

where $\text{MHSA}$ denotes the multi-head self-attention mechanism, $mQ$, $mK$, and $mV$ are the query, key, and value projections of the embeddings $mE$, and $d_k$ represents the dimensionality of the keys and queries. The output from the encoder is a high-dimensional representation that captures the attention-driven spatial relationships between different patches of the input image.

\subsection{ Efficient Time-series Transformer }
The Efficient Sequence Transformer module is designed to process sequences of NO\textsubscript{2} concentration data by employing a sparsity-enhanced self-attention mechanism to generate attention feature maps as proposed by \citet{zhou2021informer}. This mechanism, adapted from the canonical self-attention framework \cite{vaswani2017attention}, has time-complexity $O(N\text{log}(N))$ and is thus allows for more tractable compute, compared to traditional self-attention, when applied to long sequences. The authors introduce a sparsity-driven approach, \textit{ProbSparse} self-attention, which selectively computes attention weights for a subset of dominant queries, thereby reducing the computational complexity. This is based on the observation that self-attention weight distributions often exhibit sparsity, with a few key-query interactions dominating the attention mechanism. The \textit{ProbSparse} self-attention mechanism is formalized by replacing $Q$ with $Q_{\text{sparse}}$ from equation \ref{equation:attention}. Here, $Q_{\text{sparse}}$ contains only the top-$u$ queries based on a sparsity measurement $M(q_i, K)$, which quantifies the diversity of the attention distribution for each query as follows: 

\begin{equation}
M(q_i, K) = \ln\sum_{j=1}^{LK}e^{\frac{q_ik_j^T}{\sqrt{d}}} - \frac{1}{LK}\sum_{j=1}^{LK}e^{\frac{q_ik_j^T}{\sqrt{d}}}
\end{equation}

This measurement helps identify queries that contribute significantly to the attention distribution, allowing for a sparse computation of attention weights. 

The output from this module is a latent representation that captures the attention-driven temporal relationships between historical predictions of NO\textsubscript{2} concentration.

\subsection{Integration of Spatio-temporal Features via Cross Attention}

The model takes latent representations from the ViT encoder and time-series transformer to dynamically weigh the importance of spatial information from satellite imagery against the temporal patterns of predicted NO\textsubscript{2} concentrations, producing a contextually enriched feature vector \cite{gheini2021crossattention} that embodies both spatial and temporal insights. Finally, the enriched feature vector is passed through a series of fully connected layers to regress the final scalar output representing the predicted NO\textsubscript{2} concentration. The architecture is represented in figure \ref{figure:diagram}, where $m_t$ is the surface level NO\textsubscript{2} concentration prediction at timestamp $t$.

\section{Experimentation and Results}

\subsection{Data Collection and Training}

Ground-level NO\textsubscript{2} concentration data were collected from 35 monitoring stations distributed across Europe. Daily average NO concentrations\textsubscript{2} were compiled for the period October 2022 to January 2024. Corresponding Sentinel-5P satellite images were acquired for the same time period. We train the proposed model for 600 epochs on an Nvidia Tesla T4 GPU. We test the performance of the model on an unseen subset of the 35 locations that comprise our dataset, such that they are effectively out-of-distribution with respect to the training data.

\subsection{Results}
We present a comparative analysis of the metrics achieved by our proposed model on the proposed dataset, along with results achieved by \cite{deepNO2, khirwar2023geovit} on the dataset proposed by \citet{deepNO2}. For a more direct comparison, we also present an analysis of all three models on our proposed dataset. Since the dataset proposed by \citet{deepNO2} has low temporal granularity, sequences of datapoints from the same location aren't long enough for a sequence modelling module to capture useful dependencies between sequential NO\textsubscript{2} readings. Thus, the dataset proposed herein enables us to leverage sequence modeling to a more profound effect than would be possible with previously established datasets for NO\textsubscript{2} monitoring. 

This allows us to do away with the need for a Sentinel-2 image, which greatly reduces the size of the model (since Sentinel-2 images have 12 spectral channels, whereas Sentinel-5P images have only one). As is seen in table \ref{table:comparison_mae_mse_size}, our proposed model is more than an order of magnitude less in size when compared to models that use a Sentinel-2 image in conjunction with a Sentinel-5P image, while achieving or surpassing results that previous models achieve (albeit on a different dataset). We also offer a comparison of modifications of these models trained on the proposed dataset (such that they only take Sentinel-5P input), and we see that performance drops considerably, as these models do not have a time-series modeling component. In table \ref{table:comparison_mae_mse_size}, models trained on the dataset introduced by \citet{deepNO2} are marked with an asterisk (*).

\begin{table}[htbp]
\centering
\caption{Comparison of models based on MAE, MSE, and size. Best metrics are in bold.}
\begin{tabular}{lccc}
\hline
Model & MAE & MSE & Size (MB) \\
\hline
GeoViT* & 5.84 & 58.9 & 850 \\
CNN Backbone* & 6.68 & 78.4 & 1964 \\
GeoViT & 6.69 & 72.70 & 65 \\
CNN Backbone & 6.49 & 67.25 & \textbf{32} \\
GeoFormer (proposed) & \textbf{5.65} & \textbf{56.95} & 70 \\
\hline
\end{tabular}
\label{table:comparison_mae_mse_size}
\end{table}

\section{Conclusion and Future Work}

This paper presented a comprehensive approach for predicting NO\textsubscript{2} concentrations by leveraging attention between spatio-temporal features as well as a long-sequence dataset combining Sentinel-5P imagery with ground-level monitoring station readings. Future research will explore the integration of Vision Transformers with optical flow-based models \cite{guizilini2022learning} for sequence modelling, such that instead of a single image and a series of historical predictions, the model can take a series of satellite images as well as a series of historical predictions as input. Additionally, exploring the scalability of our model to other pollutants and environmental indicators could broaden the applicability of the proposed work.

\bibliographystyle{unsrtnat}
\bibliography{references}  

\appendix
\section{Dataset}
\subsection{Data Collection Process}
The following steps were taken to create the paired Sentinel-5P and NO\textsubscript{2}concentration dataset:
\begin{enumerate}
    \item \textbf{Temporal Alignment:} Sentinel-5P images were collected to align with the timestamps of the daily ground-level NO\textsubscript{2} data. To enhance temporal resolution and mitigate the effects of cloud cover and other atmospheric disturbances, images were mosaicked over rolling 10-day windows. This approach ensured that each satellite image represented an aggregate view of NO\textsubscript{2} concentrations over the 10 days preceding each ground measurement date.
    
    \item \textbf{Spatial Coverage:} For each monitoring station, a bounding box was calculated around its coordinates to define the region of interest for satellite imagery collection. This bounding box was determined based on a fixed radius from the station's location, ensuring that the satellite images encompassed the local atmospheric conditions relevant to the ground-level NO\textsubscript{2} readings.
    
    \item \textbf{Image Processing:} Satellite images were processed to match the spatial resolution and scale required for analysis. This included center cropping and resizing operations to standardize image dimensions, facilitating consistent comparison and integration with ground-level NO\textsubscript{2} data.
\end{enumerate}
\subsection{Data Samples}
\begin{figure}[ht]
\begin{center}
\includegraphics[width=0.75\columnwidth]{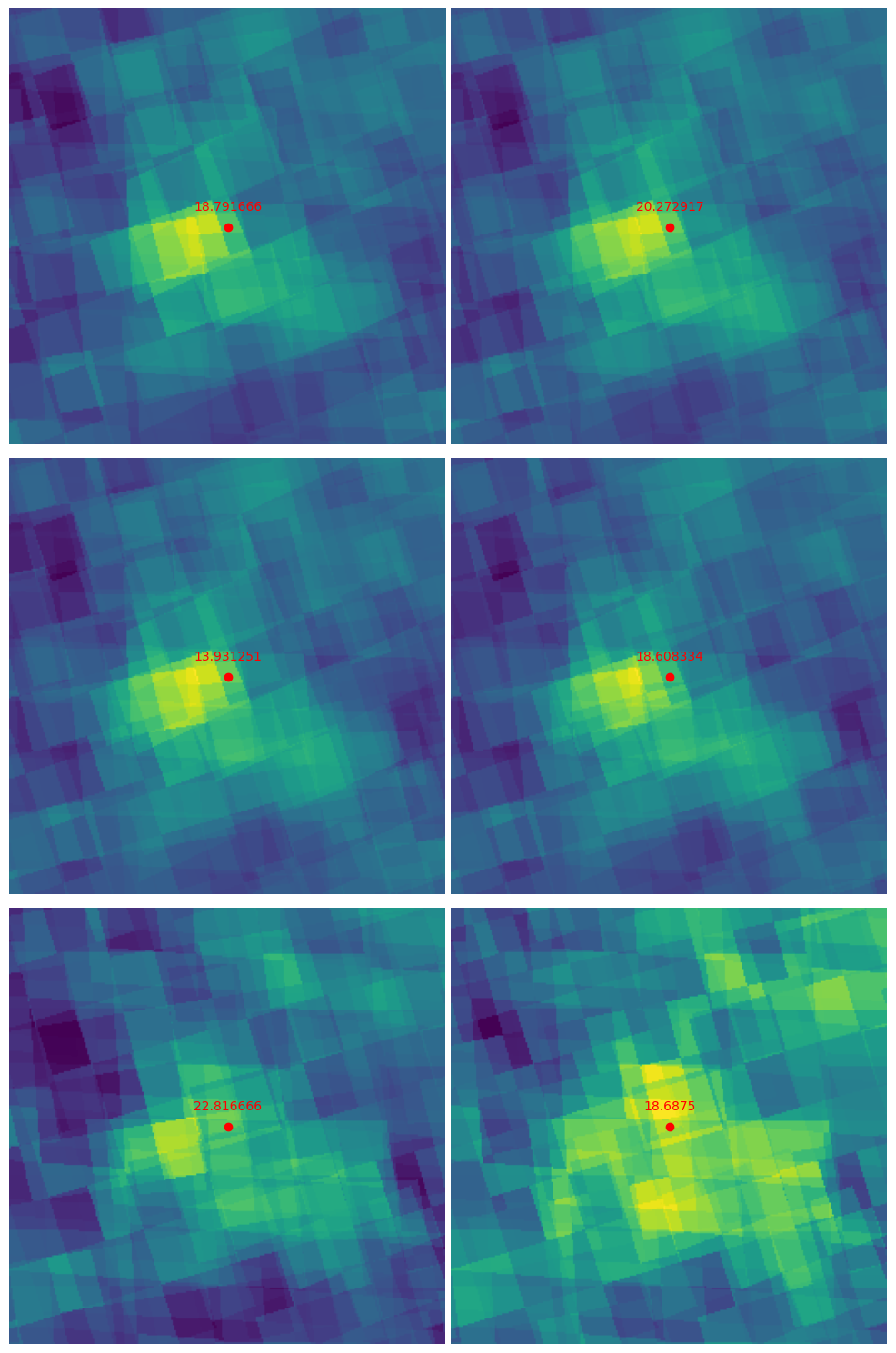}
\end{center}
\caption{Example of Sentinel-5P imagery with corresponding surface-level NO\textsubscript{2} concentrations for 6 consecutive days at the same location.}
\label{figure:dataset}
\end{figure}

\end{document}